\documentclass{article}
\usepackage{setspace}
\usepackage{amsmath}
\usepackage{algpseudocode}
\usepackage{algorithm}

\usepackage[utf8]{inputenc} 
\usepackage[T1]{fontenc}    
\usepackage{hyperref}       
\usepackage{url}            
\usepackage{booktabs}       
\usepackage{amsfonts}       
\usepackage{nicefrac}       
\usepackage{microtype}      
\usepackage{graphicx}
\usepackage{doi}
\usepackage{cite}
\usepackage[margin=1in]{geometry}
\usepackage{authblk}

\algnewcommand\algorithmicforeach{\textbf{for each}}
\algdef{S}[FOR]{ForEach}[1]{\algorithmicforeach\ #1\ \algorithmicdo}

\title{A Priori Calibration of Transient Kinetics Data via Machine Learning}

\author[1]{M. Ross Kunz}
\author[2]{Adam Yonge}
\author[1]{Rakesh Batchu}
\author[1]{Zongtang Fang}
\author[1]{Yixiao Wang}
\author[4]{Gregory Yablonsky}
\author[2]{Andrew J. Medford}
\author[1]{Rebecca Fushimi \footnote{Corresponding author: Rebecca Fushimi, rebecca.fushimi@inl.gov}}
\affil[1]{Energy $\&$ Environment Science and Technology, Idaho National Laboratory}
\affil[2]{School of Chemical $\&$ Biomolecular Engineering, Georgia Institute of Technology}
\affil[3]{Department of Electronics $\&$ Information Systems ELIS, Ghent University}
\affil[4]{Department of Energy, Environmental $\&$ Chemical Engineering, Washington University in St. Louis}

\date{\today}

\begin{document}
\maketitle

\begin{abstract}
The temporal analysis of products reactor provides a vast amount of transient kinetic information that may be used to describe a variety of chemical features including the residence time distribution, kinetic coefficients, number of active sites, and the reaction mechanism.  However, as with any measurement device, the TAP reactor signal is convoluted with noise. To reduce the uncertainty of the kinetic measurement and any derived parameters or mechanisms, proper preprocessing must be performed prior to any advanced analysis.  This preprocessing consists of baseline correction, i.e., a shift in the voltage response, and calibration, i.e., a scaling of the flux response based on prior experiments.  The current methodology of preprocessing requires significant user discretion and reliance on previous experiments that may drift over time.  Herein we use machine learning techniques combined with physical constraints to convert the raw instrument signal to chemical information.  As such, the proposed methodology demonstrates clear benefits over the traditional preprocessing in the calibration of the inert and feed mixture products without need of prior calibration experiments or heuristic input from the user. 
\end{abstract}

\textit{Keywords:} Transient Kinetics, TAP, Cheminformatics, Calibration, and Convex Optimization

\section{Introduction}
Transient kinetic investigations of catalytic materials are ideal for discriminating subtleties that may be unobservable within a steady-state experiment, leading to potentially more information about the underlying reaction mechanism \cite{bennett1982understanding}. This manuscript will focus on the kinetic information obtained by the Temporal Analysis of Products (TAP) reactor due to its well-defined operational regimes.  The generated highly resolute experimental data leads to the observed evolution of the intermediate products, thereby facilitating the expression of multiple elementary processes simultaneously \cite{morgan2017forty,gleaves1988temporal, gleaves1997tap, gleaves2010temporal}. A TAP reactor experiment consists of a series of transient responses per species that is ideal for repeatable and reproducible kinetic measurements.  However, the experimental data is imperfect in measuring the chemical signal as it may include various types of noise.  These may consist of intrinsic noise within a pulse (e.g.,  noise due to electrical signals, heater oscillations, and pulse valve drifts), noise between pulse responses (e.g., gauge error and mass spectrometer drift), or noise due to experimental setup and catalyst preparation (e.g., loading and pressure variations) \cite{roelant2007noise, constales2016advanced}. This noise directly relates to the signals within the mass spectrometer such that each measurement must be calibrated due to unique differences in baseline voltage and scaling factors for each species.  It is crucial to calibrate the outlet response appropriately to obtain accurate information about the kinetics, e.g., residence time, apparent kinetic constants, reactivities, gas concentrations, reaction rates, number of active sites, and intrinsic kinetic coefficients \cite{yablonsky2007procedure,redekop2011procedure,redekop2014elucidating, yablonsky2016rate, kunz2020probability, kunz2021data, constales2017precise}.

\begin{figure}[ht!]
	\centering
	\includegraphics[width=1\textwidth]{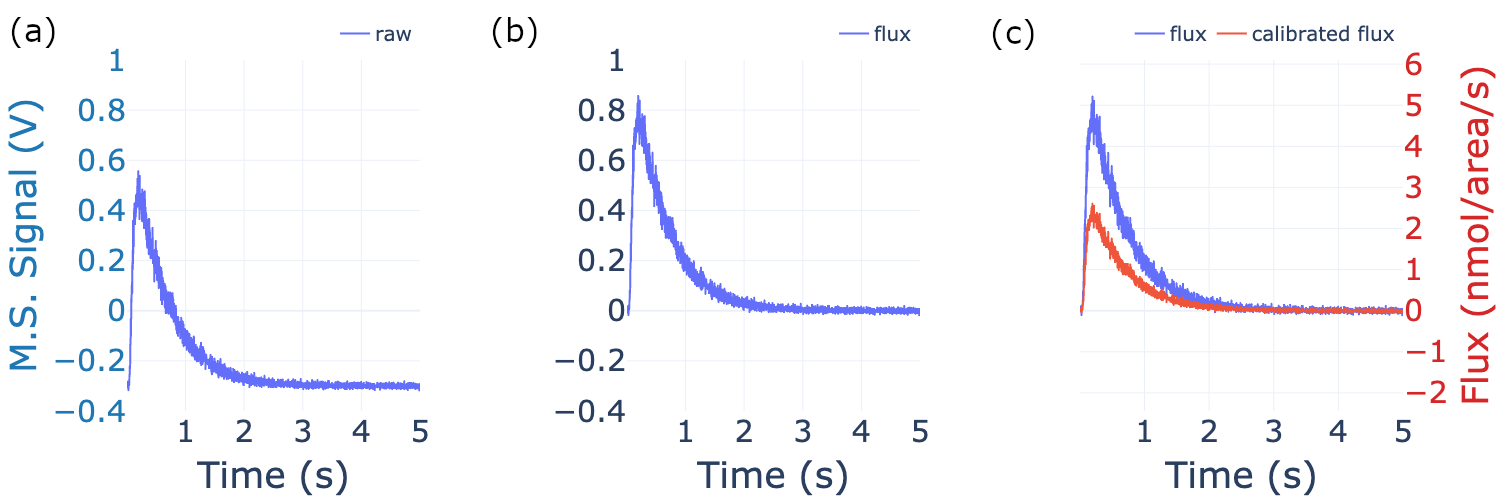}
	\caption{\label{fig:first_example} Example of the preprocessing of TAP data.  From left to right, the figure starts with a raw voltage measurement which must be baseline corrected by shifting the flux in voltage followed by calibration through scaling the flux (in red).} 
\end{figure}

The calibration of each TAP flux response differs from typical standardization techniques, i.e., subtracting the mean $(\mu)$ and dividing by the standard deviation $(\sigma)$, as this would result in a negative value for the number of molecules measured \cite{miller2018statistics, gemperline2006practical,brereton2007applied}. As such, traditional methods for calibration require each response to be baseline corrected followed by scaling by a pre-calculated mass spectrometer calibration coefficient (see Figure ~\ref{fig:first_example} as an example) \cite{kunz2018pulse,reece2017kinetic}. Current methods manually detect the baseline through visual inspection leading to potential user biases.  Additionally, significant time can be devoted to the mass spectrometer calibration by running multiple inert experiments to determine the correct scaling relationships.  However, this does not guarantee that the mass spectrometer will not drift between the inert experiment and the reaction experiment, not to mention that the mass spectrometer can drift within an experiment.   As such, misleading kinetic information may be obtained if the calibration is not properly accounted for.

A novel method is developed for treating and calibrating the data without reliance on previous inert experiments while utilizing the relationships of the reactor physics, e.g., the expected distribution of the outlet flux and statistics.  More specifically, machine learning via convex optimization with physics-based constraints is used to automatically preprocess the raw data into chemical quantities.  The baseline correction and calibration are performed through examining the relationships between the transient flux responses rather than averages or summary information, i.e., moments.  As such, it will be shown that this method allows for automatic preprocessing of TAP data for consistent kinetic estimates while saving instrument time as there is no reliance on inert experiments.

\section{Methodology}\label{section:methodology}

\subsection{The TAP reactor and standardization}
It is vital to understand the application domain prior to any analysis or preprocessing, especially if physical constraints are to be followed.  Briefly, the TAP experiment consists of an inert and reactant blend injected into the tubular micro-reactor housing a catalyst sandwiched between two inert materials in thin zone TAP reactor configuration.  Each of the products, reactants and inert species diffuse through the TAP micro-reactor and are measured by the mass spectrometer with respect to time. The inert response acts as a control variable describing the transport within the reactor as well as the total number of molecules per volt. The number of molecules within a pulse injection is typically small and is within a low-pressure regime such that only Knudsen diffusion is present in the transport.  As such, when the reactant and inert gas are injected in a one-to-one ratio, the total area of the inert response describes the number of molecules injected while the area of the reactant describes the net amount of consumption/reaction by the catalyst \cite{gleaves1997tap}.  The conversion $(\chi)$ is defined as

\begin{equation}\label{eq:conversion}
    \chi = \frac{\int F_{inert} (t)dt - \int F_{reactant} (t) dt }{\int F_{inert} (t) dt} = 1 - \frac{m^0_{reactant}}{m^0_{inert}}
\end{equation}
where $F$ represents the flux response, $t$ is time, and $m^0$ is the zeroth moment.  Further moments, e.g., $m^1$, $m^2$, and $m^3$, complete the description of the pulse by examining the mean residence time, dispersion and the kurtosis.

The flux response determines the probability distribution of measuring a molecule at a specific time where the moments of the distribution are used to determine apparent kinetic parameters \cite{gleaves1997tap, constales2017precise}. More specifically, the TAP outlet response can be approximated by the Gamma distribution:

\begin{equation}\label{eq:gamma_distribution}
    f(t; \alpha, \beta) = \frac{1}{\gamma(\alpha)\beta^\alpha} t^{\alpha - 1}e^{-t / \beta}
\end{equation}
where $t$ is time, $\alpha$ is the shape, and $\beta$ is the scale \cite{kunz2020probability}. When only diffusion is present within the reactor, an $\alpha$ of 1.5 and $\beta$ of 1/3 corresponds to the dimensionless standard diffusion curve. Using the properties of the Gamma distribution, the mean residence time $(\tau)$ is measured as $\alpha \beta = m^1 / m^0$, variance residence time $(\tau_{\sigma^2})$ as $\alpha \beta^2 = m^2 / m^0 - \tau^2$, peak residence time $(\tau_p)$ as $(\alpha - 1) \beta = argmax(flux)$, and the area normalizing coefficient $(\tau_A)$ as $\Gamma (\alpha) \beta^\alpha$.  These properties do not assume a specific reaction mechanism and when combined provide consistent explanations of the reaction with respect to each kinetic measurement. Therefore, this information can be leveraged in the standardization of the experimental data.

The TAP experiment consists of a series of transient flux responses for each of the diffused gasses from the mass spectrometer.  There are two different types of errors that can occur: the error associated within a pulse response and the error between pulse responses.  The errors within a pulse response, caused by heater oscillations, electrical signal, etc., contribute to a baseline shift in the total number of molecules, i.e., the zeroth moment of the flux $(m^0)$ \cite{roelant2007noise}. On the other hand, errors between pulse responses, caused by drift in mass spectrometer detector sensitivities etc., affect the measurements of the total number of active sites and apparent kinetics.  To account for these errors, the traditional method of calibration is done through manual centering a flux by the baseline mean $(\mu_b)$ while scaling each gas pulse series by their respective mean area $(\mu)$ of an inert experiment:

\begin{equation}\label{eq:tap_normalization}
    \boldsymbol{m}^0_C = \left(\int F_i (t) - \mu_{b,i} dt \right) / \mu
\end{equation}
where $i$ denotes each flux response in a pulse series and $\boldsymbol{m}_C^0$ denotes a vector of calibrated areas. This is similar to standard normalization practices to account for different scales and centers amongst a variety of variables. Mathematically, normalization is the process of subtracting the mean $(\mu)$ and dividing by the standard deviation $\sigma$ of a vector of responses:

\begin{equation}\label{eq:normalization}
    \boldsymbol{m}^0_N = \left( \boldsymbol{m}^0 - \mu \right) / \sigma
\end{equation}

where $\boldsymbol{m}^0_N$ is a vector of normalized areas.  Equations~\ref{eq:normalization} and \ref{eq:tap_normalization} both have the same goal of standardizing the data, but the shift in Equation~\ref{eq:tap_normalization} is constrained such that the number of molecules measured is not negative.  Additionally, the scaling factor differs by using a first or a second order moment.  However, if the distribution of the $\boldsymbol{m}^0$ is skewed, the first and second order moment have a linear relationship and hence can be used interchangeably \cite{casella2002statistical}.  As such, it is the goal to show the relation between these two equations such that advanced data analysis may be applied to the traditional calibration.  This will be done through smoothing the flux to account for the noise within a flux, baseline correction to account for the mean shift or centering, and calibration to account for the scaling factor between species.

\subsection{Smoothing the flux} \label{subsection:smooth}

First, to separate the chemical signal from the random noise within a flux, a univariate cubic spline approximation is applied \cite{de1978practical}. This enables robust estimation of the distributional parameters, such as the peak residence time, while ensuring that the flux is neither under/over-represented.   While this step is minor in terms of preprocessing, it ensures that the calibration is only determined by the chemical signal and not the noise. This reduction in noise within a flux is critical in obtaining accurate estimates of the residence time distributional properties as the heteroscedastic nature of the noise will affect the measurement of the moments \cite{roelant2007noise}.

\subsection{Baseline correction} \label{subsection:baseline}

In accordance with the probability distribution and the calibration requirements, the baseline of the flux response must be zero.  However, in an experimental setting, the baseline may shift due to within pulse variation and the baseline may not be reached for chemical species that have longer elution time than the collection time.  To address this issue, the baseline can be estimated using the shape and scale parameters of the Gamma distribution of a flux.  The shape of the Gamma distribution affects the outlet flux by time, i.e., time of position in the reactor, while the scale affects the kinetic rates \cite{kunz2020probability}. As such, the same shape parameter should be used in either case of pure diffusion or reaction occurring in the outlet flux.  The peak residence time, i.e., the mode, property of the distribution is used for the estimation of the distribution parameters as it will not be affected by shifts in the baseline.  Assuming a shape parameter of an ideal reactor at 1.5, then the scale of a flux may be estimated, using the mode property of the Gamma distribution, as two times the peak residence time, i.e., $\hat{\beta} = 2\tau_p$. Once the scale is estimated, the probability of the distribution at the end of the experiment time may be determined by Equation~\ref{eq:gamma_distribution}. The tail of the flux can then be shifted to match the probability of the Gamma distribution providing a similar shift to the normalization procedure in Equation~\ref{eq:normalization}.  Algorithm~\ref{alg:baseline} describes the steps required to perform baseline correction based on the Gamma distribution.

\begin{algorithm}
\caption{Baseline correction via distributional assumptions. }\label{alg:baseline}
\begin{algorithmic}[1]
\Require Let $F$ represent the gas flux, $t_{end}$  denote the time at the end of the flux, $f(\cdot)$  denotes the Gamma distribution defined in Equation~\ref{eq:gamma_distribution}, and $F[t_{end} ]$ denotes the voltage of the flux at the end of the experiment time.
\State $\tau_p = argmax (F)$
\State $\Gamma_{tail} = f(t_{end}; 1.5, 2 \tau_p)$
\State $\bar{F} = F - F[t_{end}] + \Gamma_{tail}$
\end{algorithmic}
\end{algorithm}

\subsection{A priori calibration between gas species using moments} \label{subsection:momentCalibration}

Ideally, the scale of the voltage response given by the mass spectrometer would measure the exact number of molecules for each response.  However, the differences in voltage scale between gases is dictated by multiple attributes which may include the ionization efficiency, gain of the electron multiplier, and mass differences based on instrument drift \cite{ko1980reactions}. It is necessary to determine the appropriate scale, based on the known feed mixture blend amounts, to measure the kinetic coefficients, gas concentrations, and reaction rates.  These scaling coefficients, i.e., calibration coefficients $(\zeta)$, are traditionally calculated within an inert experiment where the mean $\bar{m}_g^0$ of the reactants/products are scaled to the mean inert $\bar{m}_I^0$, i.e., $\bar{m}_g^0 = \zeta \bar{m}_I^0$, where $\bar{m}$ denotes the mean moment.  If the baseline relationship between the two gases is exactly zero, i.e., the baseline is appropriately determined for both species, the calibration coefficient $\zeta$ can be directly solved through the use of an Ordinary Least Squares (OLS) model where the scaling relationship is due to the standard deviation as in Equation~\ref{eq:normalization}. Note that the scaling factor $(\mu$ or $\sigma)$ in Equations~\ref{eq:tap_normalization} or \ref{eq:normalization} can be used interchangeably due to the linear relationship between the mean and variance of the Gamma distribution, i.e., $\tau_{\sigma^2}/\tau = \beta$.  The use of a linear model, rather than the mean estimate, has additional benefits as the intercept can be verified if it is exactly zero and robust modifications may be used to account for outgassing within the series, i.e., an additional pulsed amount of gas within the reactor at an unexpected time \cite{andrews1974robust}.  

In application to a reaction experiment, changes in the mass spectrometer measurement are assumed to be fixed when moving from the previous inert experiment.  However, this is rarely the case as the instrument will drift in time. Constant inert calibration is impractical as this would require a complete work stop to exchange the catalytic material to an inert bed in the micro-reactor and hence potentially change the experimental conditions.  However, the calibration coefficient can be extended to a reaction setting using the information gained by the Gamma distribution.  More specifically, the distribution normalizing coefficient $\boldsymbol{\tau}_A$ estimates the effect of conversion based on the residence time distribution properties.  As such, the $\boldsymbol{m}^0_g$ is a linear combination of the area due to the conversion and the instrument drift:

\begin{equation}\label{eq:linear_moments}
    \boldsymbol{m}^0_g = \mu + \zeta_1 \boldsymbol{\tau}_A + \zeta_2 \boldsymbol{m}_I^0
\end{equation}
where $\mu$, $\zeta_1$, $\zeta_2$ are estimated through application of an OLS model.  The value of $\mu$ is the intercept and hence determines the shift between each species.  If the baseline was determined prior to Equation~\ref{eq:linear_moments}, the the value of $\mu$ should be zero.  The value of $\zeta_1$ indicates the scaling relation to the $m^0_g$ and the distribution normalizing coefficient.  The size of $\zeta_1$ is based on the amount of conversion.  The last term $\zeta_2$, is the regression coefficient describing the scale from the inert to the gas species, i.e., accounting for the mass spectrometer differences between each gas species.  Thus, a calibration coefficient from the inert to the reactant/product $m^0_g$ can be obtained even in the presence of reaction. 

After calibration, the following $m^0$ relationships should hold between the reactants (r), products (p), and the inert (I):

\begin{align*}
    \mbox{If the reaction is reversible:}& \\
    & m^0_r = m^0_I \mbox{ and } \tilde{m}^1_r > \tilde{m}_I^1 \\
    \mbox{If the reaction is irreversible:}& \\
    & m^0_r < m^0_I, \tilde{m}^1_r < \tilde{m}_I^1 \mbox{ and }  m^0_r + m^0_p \leq m^0_I
\end{align*}
where $\tilde{m}^1$ is the area normalized first moment.  This information assumes an 50/50 blend pulse injection between the inert and the reactant.  However, this still applies to unequal blends, but a scaling coefficient based on the blend amounts must be applied, e.g., if reversible, then the $m^0$ amounts per inert and reactant correspond to the blend amount. The first relationship states that if the normalized first moment of the reactant is greater than the inert, then the reaction is reversible, and the calibration coefficient can be set such that the area of the reactant equals the area of the inert. The second relationship states that the area of the reactant must be less than the area of the inert based on the amount of conversion.  If the normalized first moments are equal, then no reaction has occurred.  The final relationship describes the balance between the number of molecules injected and the combination of the reactants and products.  The less than or equal to sign allows flexibility in the event there is accumulation of a species on the surface of the catalyst.  Algorithm~\ref{alg:momentCalibration} describes the steps required for calibration within a reaction experiment using momentary information.

\begin{algorithm}
\caption{Calibration scaling via distributional assumptions.}\label{alg:momentCalibration}
\begin{algorithmic}[1]
\Require Let $\bar{F}$ represent the gas flux that is baseline corrected using Algorithm~\ref{alg:baseline}, $I$ denote the inert gas, and $g$ denote a reactant/product gas.
\ForEach{$i^{th}$ pulse}
    \State $m^0_{I,i} = \int \bar{F}_{I,i} (t) dt$
    \State $m^0_{g,i} = \int \bar{F}_{g,i} (t) dt$
    \State $\tau_{A,i} = \Gamma(\alpha_i)\beta^{\alpha_i}_i \mbox{ of } \bar{F}_{g,i}$
\EndFor
\State Solve $\boldsymbol{m}^0_g = \mu + \zeta_1 \boldsymbol{\tau}_A + \zeta_2 \boldsymbol{m}_I^0$ via OLS or robust OLS.
\end{algorithmic}
\end{algorithm}

\subsection{A priori calibration between gas species using transient information and machine learning} \label{subsection:tcco}

Given the moment-based relationships between the gas species, the relationship between the transient responses may be leveraged in a similar fashion to Equation~\ref{eq:linear_moments}. The goal then is to compare the gas to the inert species per pulse to determine the appropriate calibration coefficient for an irreversible reactant and each product within a reaction experiment.  First, consider an irreversible reactant and the reactant flux has been transformed to have the same velocity as the inert flux via Graham’s Law.  Based on the $m^0$ relationship, the reactant flux must have a positive scaling relationship to the inert flux, the difference between the calibrated reactant flux and the inert flux must be greater than or equal to zero when a product/surface species is formed, and the area of the reactant flux must be less than the inert flux.  Numerically, the calibration coefficient can be solved as a linear model between the inert and the reactant with the following constraints:

\begin{align}
    &\min_b \sum^N_{i=1} \left( F_{I,i} - F_{r,i}b \right)^2 \label{eq:tcco1}\\
    \mbox{such that: } &b \geq 0, F_{I,i} - F_{r,i}b \geq 0, \mbox{ and } m^0_I \geq m^0_r \label{eq:tcco2}
\end{align}
where $b$ is the calibration coefficient from the reactant species to the inert species. The objective function and the first constraint consist of a variation of OLS called non-negative least squares, the second constraint ensures there is not a negative number of molecules measured in the residuals, and the last constraint ensures that the area of the reactant flux is less than the inert flux area.  The combination of the objective function and the constraints will be denoted as Temporal Convex Constrained Optimization (TCCO) for ease of referencing throughout the paper.  

The application of TCCO does not have to be restricted to the relationship between the inert and a reactant gas species.  For example, a series of inert flux may be calibrated to a single inert reference sample such that all fluxes are calibrated to one pulse injection.  In general, TCCO can be applied to several reactants and products or even mass spectrometer fragmentations but requires the alteration of Equations~\ref{eq:tcco1}-\ref{eq:tcco2} to a multiple linear regression form.  Let $\boldsymbol{y}$ be a vector of the flux in which to calibrate to, i.e., the Dependent Variable (DV) typically the inert flux, $\boldsymbol{X}$ be a matrix of flux responses, i.e., the set of Independent Variables (IV), and $\boldsymbol{res}$ denote the residuals, then

\begin{align}
    &\min_b || \boldsymbol{y} - \boldsymbol{Xb}||^2_2 \label{eq:tccoMLR1}\\
    \mbox{such that: } &\boldsymbol{b} \geq 0, \boldsymbol{res} \geq 0, \mbox{ and } m^0_y \geq m^0_X \label{eq:tccMLR2}
\end{align}
where $m^0_X$ is the sum of the zeroth moments for each variable in $\boldsymbol{X}$. Equations~\ref{eq:tccoMLR1}-\ref{eq:tccMLR2} are convex in optimization and can be solved using open-source solvers \cite{agrawal2018rewriting,diamond2016cvxpy}.

\subsection{Transient Error Analysis for Kinetic measurements} \label{subsection:TEAK}

The smoothing, baseline correction, and calibration coefficient calculation are combined to form the Transient Error Analysis for Kinetic (TEAK) measurements algorithm.  For ease of use, the complete workflow is given in Algorithm~\ref{alg:TEAK}.  Software implementation of TEAK as well as commonly used TAP analysis techniques can be found at \url{https://github.com/IdahoLabResearch/tapsap}.

\begin{algorithm}
\caption{Transient Error Analysis for Kinetic measurements (TEAK).}\label{alg:TEAK}
\begin{algorithmic}[1]
\Require Let $F$ represent the gas flux, $smooth$ represent the application of the spline approximation in Section~\ref{subsection:smooth}, $I$ denote the inert gas, and $g$ denote a reactant/product gas.
\ForEach{$i^{th}$ pulse}
    \State $\hat{F}_{g,i} = smooth(F_i)$
    \State $\bar{F}_{g,i} = baseline(\hat{F}_{g,i})$
    \State $\tilde{F}_{g,i} = tcco(\hat{F}_{I,i}, \hat{F}_{g,i})$
\EndFor
\end{algorithmic}
\end{algorithm}

\section{Experimental Data} \label{section:experimentalData}

\subsection{Inert experiment with varying temperature} \label{subsection:temperature}

A series pulse response data of argon was obtained from the Temporal Analysis of Products reactor, TAP-3E. The tubular micro-reactor housing in TAP-3E is of $0.00385 m$ internal diameter and $0.0564 m$ in length. In Knudsen conditions, argon is pulsed over an inert quartz bed $(250-300 \mu m)$ at five different temperatures between 100 to $500 ^{\circ}C$ with $100^{\circ}C$ interval. At each of the temperature, two identical experiments are conducted for reproducibility. Each experiment is recorded for 100 pulses with a collection time of $5 s$ and pulse spacing of $5.1 s$. Argon is monitored at $m/z = 40$ Atomic Mass Unit (AMU) from a quadrupole mass spectrometer.

\subsection{Inert experiment with calibration} \label{subsection:inertCalibration}

The inert experiments were performed in TAP by pulsing carbon dioxide and argon (volume ratio with 50:50) over quartz at room temperature. An electronic pulse width of $135 \mu s$ was sent to the valve every $5.1 s$ for a total of 1000 pulses. At each pulse a different atomic mass unit (AMU) was monitored cycling between 40, 44, 28 16 and 12 to record the pulse response for argon, carbon dioxide, the fragmentation of carbon dioxide on mass 16 and 12, respectively. 

\subsection{Reaction experiment} \label{subsection:dataReaction}
The Strong Electrostatic Adsorption (SEA) method was used for the synthesis of $1.0 wt\%$ $Pt/SiO_2$ catalyst for the oxidation experiments.  A commercial silica (AEROSIL OX50, $50 m^2/g$) from EVONIK was chosen as the support and a precursor of tetraamine platinum (II) hydroxide $(Pt(NH_3)4(OH)2$, $99\%$, from Aldrich) was employed to deposit the metal. The precursor was dissolved in deionized (DI) water and the initial pH was adjusted to 11.5 with NaOH. Silica was added to the solution and the contents were shaken for one hour.  The resulting mixture was washed with DI water, filtered, and dried overnight under vacuum.  The material was pressed and sieved, retaining the $250 - 300  \mu m$ fraction.  Next, the catalyst was pretreated ex-situ in $50\%$ oxygen and argon flow $(30 mL/min)$ at $400^{\circ}C$ for $30 min$ followed by reduction in $4\%$ hydrogen and argon flow $(50 mL/min)$ at $400^{\circ}C$ for one hour. The ex-situ oxidation and reduction was performed with three cycles.

Approximately $15.6 mg$ of pretreated catalyst with the particle size of $250 - 300 \mu m$ was loaded between two zones of the same particle size quartz sand (Sigma Aldrich).  
The total length of the reactor was $0.0564 m$, with a catalyst zone of $0.002 m$, and a cross sectional area of $1.256 \times 10^{-5}  m^2$.  
The TAP reactor was evacuated at $300^{\circ}C$ to a pressure of $1 \times 10^{-7} torr$ and the catalyst was subjected to at least three cycles of alternating pulses of 200 pulses of carbon monoxide and argon and 200 pulses of oxygen and argon to activate the platinum and reach a reproducible starting point for pulsing experiments. Prior to oxygen adsorption, the catalyst was again reduced at $300^{\circ}C$ by introducing a sequence of $50\%$ carbon monoxide and argon pulses until no carbon dioxide formation was detected. The TAP reactor was subsequently heated to $500^{\circ}C$ and kept for $30 min$ to remove adsorbed carbon monoxide and then cooled to the desired temperature for testing oxidation. The adsorption of oxygen on the catalyst was recorded in separate experiments by pulsing a 1:1 oxygen and argon mixture at $300^{\circ}C$ with different pulsing intervals of 2.0, 2.5, 3.0, 3.5, and $4.0 s$. The time evolution of three mass fragments was followed, namely argon (AMU 40), oxygen (AMU 32), and carbon dioxide (AMU 44). There was no carbon dioxide production detected at the beginning of each oxidation experiment. 

\section{Results and Discussion} \label{section:RD}

To thoroughly test the TEAK preprocessing methodology, several different cases are applied: simulated data with a known outlet flux calibration coefficient, an inert experiment where argon is exposed to multiple temperatures, an inert experiment where two different gases are calibrated to one another, and finally a reaction experiment where oxygen interacts with a platinum catalyst. Each simulation/experiment is selected to show the benefits of TEAK in an incremental fashion.  The simulated data shows the benefits of calibration from flux to flux using TCCO, the inert experiment with temperature variation gives an example of TCCO over multiple data sets of a similar nature, the inert experiment with calibration shows how TEAK deals with non-linear drift, and the reaction experiment shows how TEAK can achieve consistent measurements of the conversion without prior calibration experiments.

\subsection{ Application to simulated data}

\begin{figure}[ht!]
	\centering
	\includegraphics[width=1\textwidth]{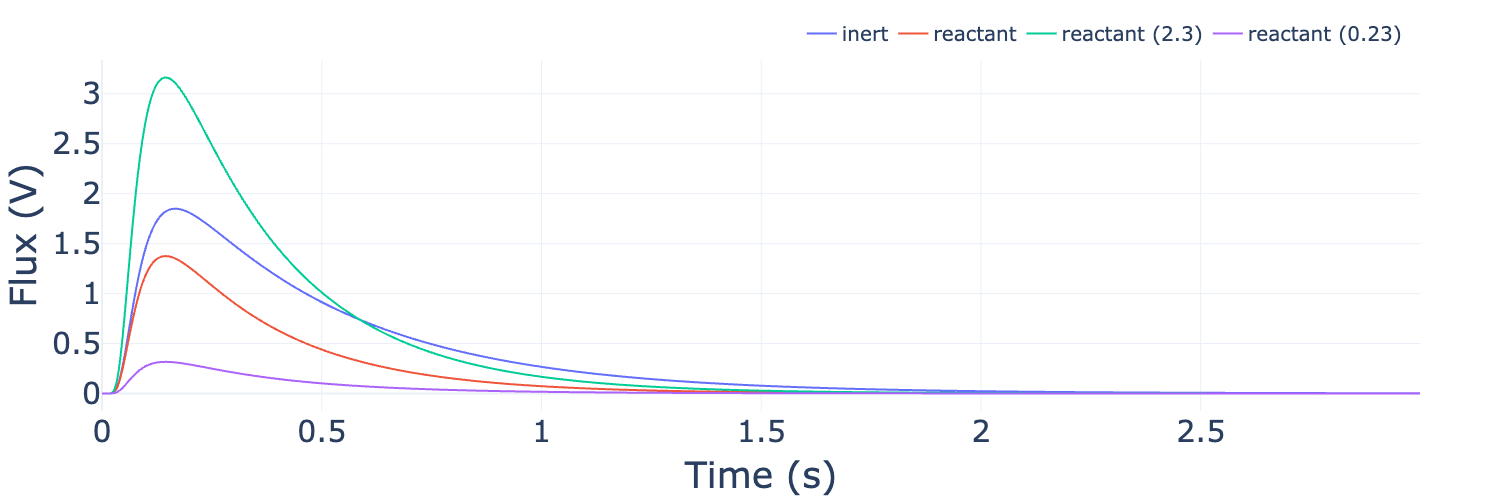}
	\caption{\label{fig:simulated} Simulated reaction data that is scaled arbitrarily to test the constrained convex optimization in Section 2.5.} 
\end{figure}

Simulated data was generated to produce a chemical signal with arbitrary scaling values.  The simulated reactor consists of a thin-zone setup with the total length equal to $1 cm$, a bed porosity of 0.5, and a diffusion coefficient of $0.5 cm^2/s$. A total of $1 mol$ was injected into the reactor and measured over three seconds. The reaction consists of a single elementary step where a reactant gas irreversibly interacts with a catalyst at the rate of $1.15 mol/s$.  After which, the reactant outlet flux was scaled arbitrarily by values of 2.3 and 0.23 to simulate the mass spectrometer measuring the reactant at a non-ideal scale as shown in Figure~\ref{fig:simulated}.

The data generated is ideal for validating potential calibration coefficients for TEAK as it has a known calibration coefficient and no smoothing nor baseline correction is required.  Only irreversible reaction is considered in the simulation as the reversible reaction would return the same number of molecules as the inert flux.  Application of TCCO assumes the DV is the inert flux while the IV only consists of the scaled reactant flux (2.3 or 0.23). TCCO applied to the inert and scaled reactant flux determined the calibration coefficient within $1 \times 10^{-7}$ of the true calibration coefficients (2.3 or 0.23).

\subsection{Application to the inert experiment with varying temperature}

The inert experiment with varying temperature consists of measuring pulsed argon within an inert experiment over several temperatures.  The objective of this example is to verify that given clean data, the TEAK methodology applied from flux to flux is consistent with the traditional series of moment-based calibration.  First, each individual temperature was pre-processed independently where the traditional method baseline corrected each flux by the mean of the flux tail (from 4.5 to $5 s$).  The zeroth moments produced by the traditional method were then compared to the data being baseline corrected using Algorithm 1 followed by using TCCO within the flux series.  More explicitly, a single pulse index was selected arbitrarily (3rd pulse) as the DV and TCCO was applied to each other flux in the series as the IV independently.  

\begin{figure}[ht!]
	\centering
	\includegraphics[width=1\textwidth]{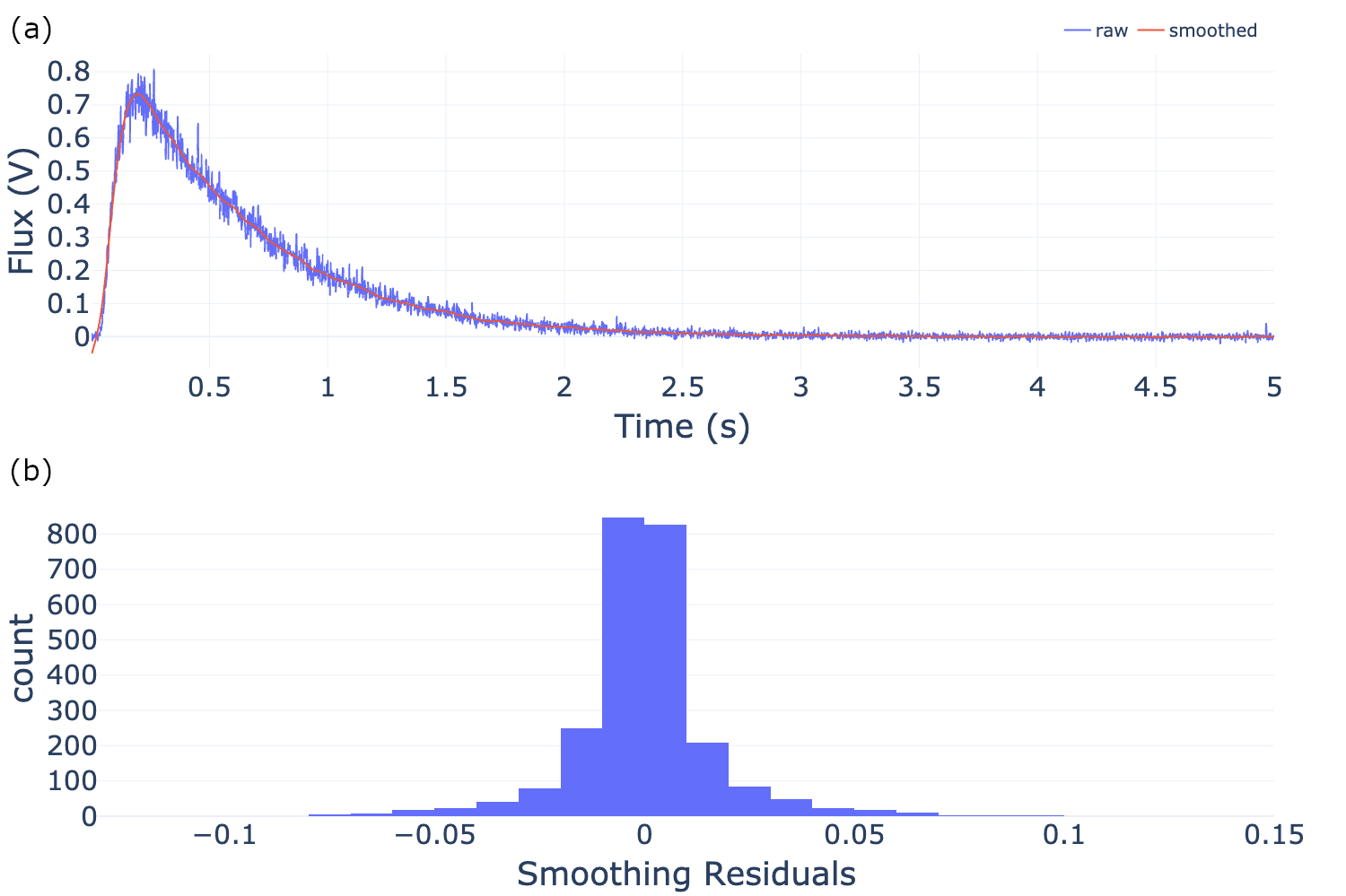}
	\caption{\label{fig:smooth} A single spline smoothed argon response from the inert data with varying temperature at $100^{\circ}C$. Subfigure (a) describes the relationship between the noisy outlet response and the estimated trend.  Subfigure (b) describes the obtained residual noise.} 
\end{figure}

Prior to the application of TEAK, each of the flux were smoothed independently via the cubic spline given in Section~\ref{subsection:smooth}. This allows for a better estimate of the peak residence time used in baseline correction and calibration to the signal rather than the noise in TCCO. For example, Figure~\ref{fig:smooth} displays the smoothed flux where the peak height and flux shape does not significantly differ from that in the flux response.  Additionally, the residuals of the smoothing produces a Gaussian distribution consistent with previous studies \cite{roelant2007noise}.

\begin{figure}[ht!]
	\centering
	\includegraphics[width=1\textwidth]{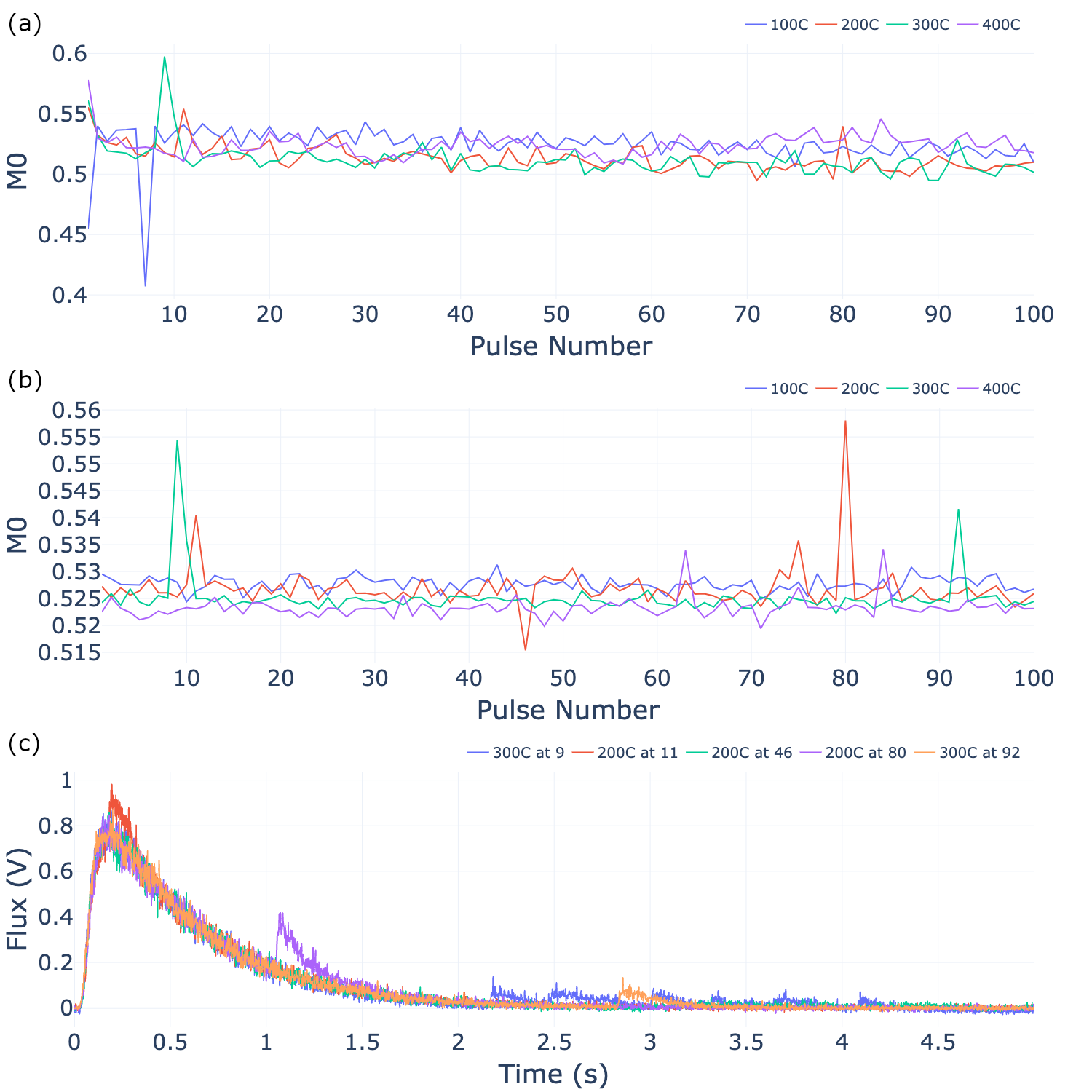}
	\caption{\label{fig:temperatureData} Comparison of the traditional calibration in subfigure (a) and TEAK in subfigure (b) applied to the inert temperature varying data.  Subfigure (c) gives examples of outgassing which affects the zeroth moment measurements.} 
\end{figure}

The comparison between the $m^0$ measurements after calibration can be seen in Figure~\ref{fig:temperatureData}.  The zeroth moments of the traditional method exhibits a minor slope over the pulse number while there is no drift within the TEAK results.  It would at first appear that TEAK exhibits poor measurements in select pulse numbers, e.g., 9th, 16th, 80th, etc.  However, upon closer inspection of the individual flux, each of the respective flux contains outgassing. These moments with outgassing are easily identified using TEAK where a user may exclude these points from the kinetic analysis.  Comparison of the mean zeroth moments show relatively similar values of 0.518 and 0.525 for the traditional and TEAK methods.  However, TEAK significantly reduces the variance of the zeroth moments, by a factor of 32 times, after accounting for the outgassing effects.

\subsection{Application to the inert experiment with calibration}

The inert experiment with calibration shows how TEAK can account for non-linear drift between pulse measurements and how to calibrate two different gas species (argon and carbon dioxide).  Additional fragmentations of the carbon dioxide were inspected to show evidence of the Gamma distribution relationship between the mean and variance residence time.

\begin{figure}[ht!]
	\centering
	\includegraphics[width=1\textwidth]{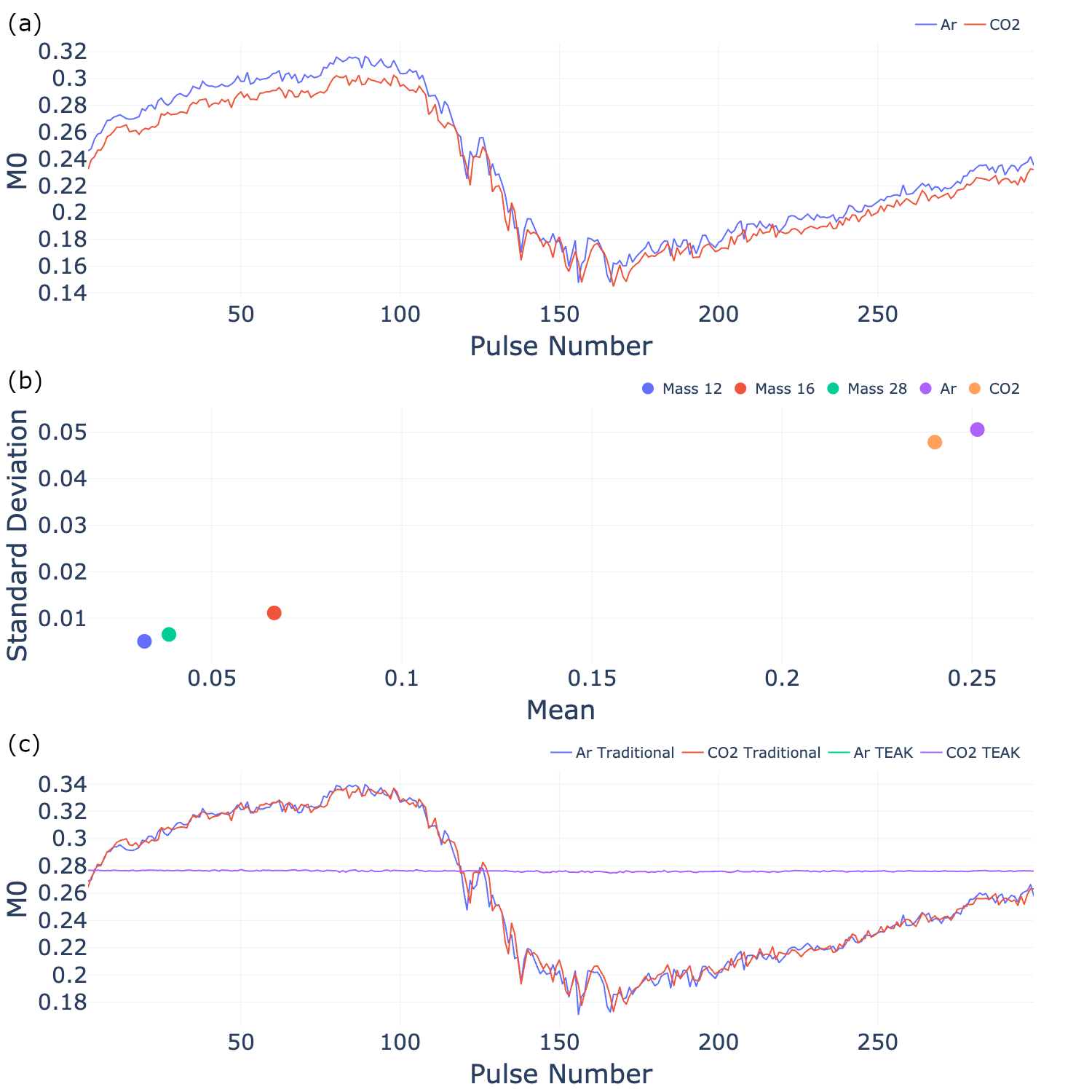}
	\caption{\label{fig:nonlinear} The raw inert reactor information for each mass collected by the mass spectrometer.  Subfigure (a) displays each uncorrected $m^0$ value of argon and carbon dioxide with respect to the pulse index number. Subfigure (b) denotes the relationship between the mean and standard deviation of the argon, carbon dioxide, and fragmentations. Subfigure (c) displays the corrected $m^0$ for the traditional calibration and TEAK.  The argon and carbon dioxide values of TEAK are visually indiscernible when compared to the same scale of the traditional calibration. } 
\end{figure}

First, each flux was baseline corrected individually, from the flux tail using the traditional method and from Algorithm~\ref{alg:baseline} in TEAK. The zeroth moments of each mass were measured for determining the calibration coefficient in the traditional method.  Figure~\ref{fig:nonlinear} (a) displays the zeroth moments prior to calibration. A sinusoidal trend over the $m^0$ responses can be detected through visual inspection. This trend exposes a weakness in the traditional method of calibration as different means of the $m^0$ may be extracted depending on when the experiment was conducted with respect to the pulse index. Figure~\ref{fig:nonlinear} (b) compares the mean and standard deviation of the zeroth moments.  There exists a linear trend of the mean and standard deviation with a slope of 0.2.  This linear relationship indicates a skewed distribution, such as a Gamma distribution, within the mass spectrometer as indicated in Section~\ref{subsection:momentCalibration}.

After baseline correction, the calibration coefficient for the traditional method was calculated as the ratio of the mean zeroth moments over the whole series.  The TEAK method applies two forms of calibration. First, TCCO was applied to each argon flux using the 5th argon pulse as the DV. The choice of the DV pulse number will change the overall $m^0$ of each calibrated flux, but the relationship between the DVs and IVs after calibration is consistent. Additionally, the choice of a DV should not include any outgassing effects. Second, TCCO was applied to each corresponding pulse number of carbon dioxide as the IV where the argon flux was the DV. Figure~\ref{fig:nonlinear} (c) compares the $m^0$ measurements for argon and carbon dioxide obtained by the traditional method and TEAK.  The argon TEAK pulse is difficult to discern as it almost exactly corresponds to the carbon dioxide series.  TEAK removes the non-linear drift as well as reduces the mean square error of the zeroth moment difference between argon and carbon dioxide by nearly 3 orders of magnitude ($9.5 \times 10^{-3}$ compared to $1.1 \times 10^{-5}$).

\subsection{Application to the reaction experiment}

The final example consists of an irreversible reaction of oxygen interacting with a platinum catalyst. This experiment is particularly difficult for the traditional calibration as there exists mass spectrometer drift from pulse to pulse, the calibration coefficient from the inert experiment is not an exact description of the mass relation, and there are many instances of outgassing.  Additionally, with reaction, the $m^0$ response will change the mean and variance of each mass in the reaction depending on the amount of conversion.  In the traditional calibration method, it is assumed that the mass spectrometer variation within an inert experiment may be used to dictate the drift in the reaction experiment.  TEAK avoids this assumption by performing the calibration with respect to each individual flux response using TCCO.

In preparation of the zeroth moment values, the traditional method preprocessing consisted of baseline correcting the flux via the tail of the flux and scaling oxygen to argon through a calibration coefficient of a prior inert experiment.  The TEAK preprocessing workflow consisted of smoothing the data, baseline correction via Algorithm~\ref{alg:baseline}, application of TCCO to argon using a reference argon flux, and application of TCCO from the argon flux (the DV) to the oxygen flux (the IV) on a pulse-by-pulse basis.

\begin{figure}[ht!]
	\centering
	\includegraphics[width=1\textwidth]{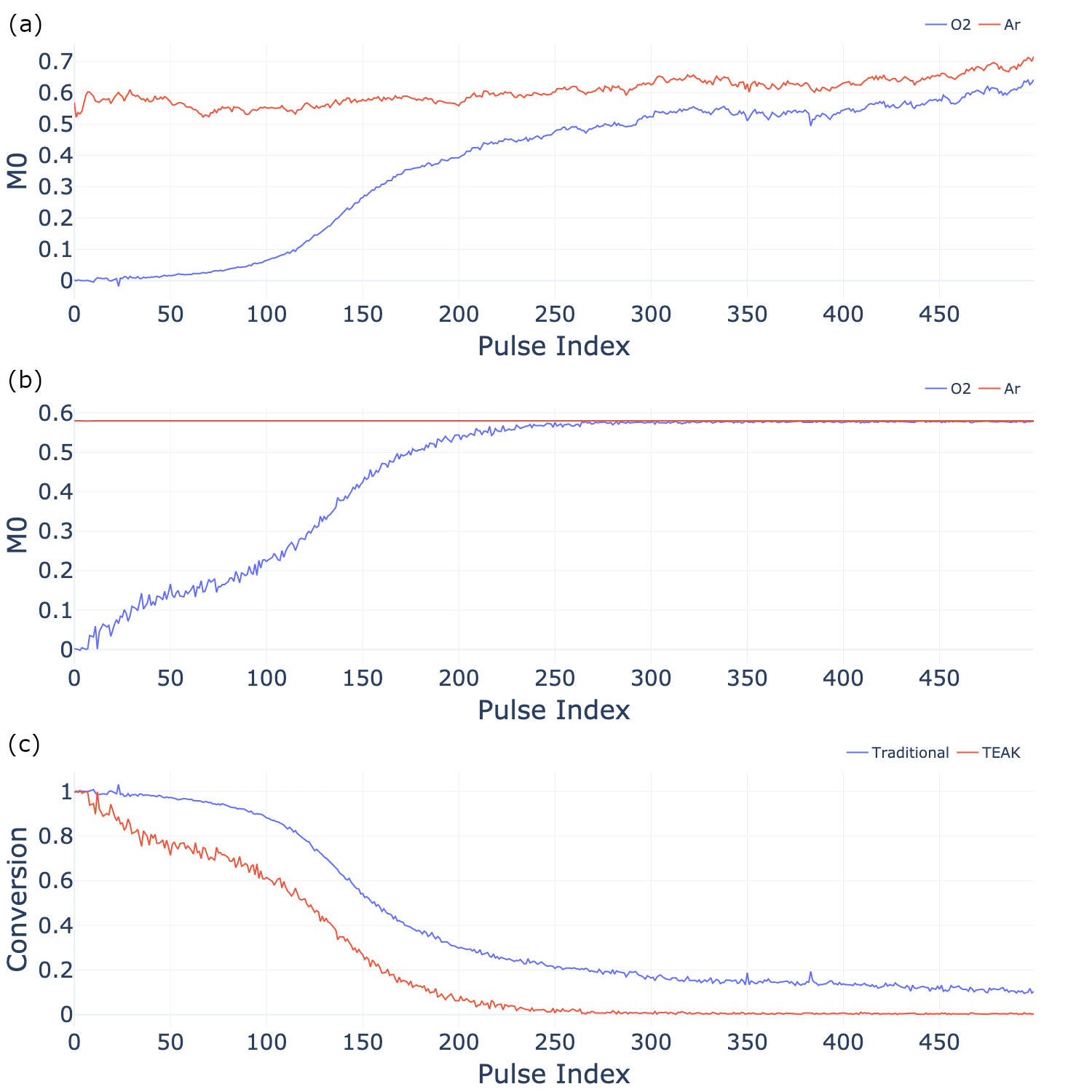}
	\caption{\label{fig:pt} Zeroth moment values obtained from the traditional calibration in subfigure (a) compared to TEAK in subfigure (b). Subfigure (c) compares the two different conversion values obtained for each method.} 
\end{figure}

The resulting zeroth moments of each method are given in Figure~\ref{fig:pt} (a) and (b) after removal of the outgassing pulses.  The removal of the outgassing effects can be accomplished automatically using a simple moving window $t$-test or through time series anomaly detection methods \cite{ljung1993outlier}.  However, it is recommended to confirm the outgassing through visual inspection prior to removal.  The argon zeroth moment of the traditional method exhibits drift from pulse to pulse while the oxygen, after approximately 300 pulses, appears to follow the same drift pattern. It is difficult to determine at what point the variation over a series of pulses were due to drift or conversion.  On the contrary, the TEAK argon zeroth moments are consistent across the pulse number while the oxygen values converge to the argon series as the pulse number increases. Significant differences in the measured zeroth moments exist early in the pulse index (less than 150) which will affect the interpretation of the characterization. Figure~\ref{fig:pt} (c) displays the conversion values for each method, i.e., $1 - m^0_{O_2} / m^0_{Ar}$. The TEAK method was able to capture the conversion values early in the pulse number where the noise dominates the signal.  Additionally, TEAK showed that the oxygen is no longer being consumed at the 300th pulse number while the traditional method suggests that reaction may still be occurring.  The zeroth moment measurement can be considered the foundational characterization of the experiment as it relates to the conversion, apparent kinetics, Damk{\"o}hler number, and measurement of the total number of active sites \cite{constales2017precise,constales2019methods}. Furthermore, when examining transient kinetic responses, the reaction rate and gas concentration measurements will be affected by the scale of the flux.  More specifically, the reaction rate is determined as a function of the difference of the inert diffusion and the reactant flux response, i.e., $R = f(I-F)$ \cite{yablonsky2007procedure, redekop2011procedure, kunz2018pulse, kunz2020probability}. When the inert is improperly scaled, the transient rate will be over/underestimated with respect to the concentration and may lead to incorrect conclusions regarding the chemical kinetics.

\section{Conclusion}

This manuscript has demonstrated the use of residence time distribution properties and machine learning through convex optimization with physical constraints to deconvolute a noisy signal into a chemical signature.  This type of analysis was enabled through the repeatable and reproducible measurements of the TAP reactor between each gas/surface interaction and over a series of pulse responses.  The developed methodology allows for separation of the noise due to external environmental factors, detrending the series of outlet responses due to drift within the system, and mass spectrometer calibration coefficient estimates without prior inert experiments.  

The proposed methodology was verified through analysis of inert and reaction experiments with a range of conditions.  As a result, more consistent estimates of kinetic measurements are obtained over the traditional calibration methodology.  Additionally, the proposed methodology does not rely on user input for standardization of the data and hence has less potential for user biases while creating a streamlined process to convert instrument measurements to kinetic information.

\newpage

\bibliographystyle{ieeetr}
\bibliography{tap.bib}
\end{document}